\documentclass[lettersize,journal]{IEEEtran}
\usepackage{amsmath,amsfonts}
\usepackage{algorithmic}
\usepackage{algorithm}
\usepackage{array}
\usepackage[caption=false,font=normalsize,labelfont=sf,textfont=sf]{subfig}
\usepackage{textcomp}
\usepackage{booktabs}
\usepackage{stfloats}
\usepackage{url}
\usepackage{verbatim}
\usepackage{graphicx}
\usepackage{cite}
\usepackage{caption}
\usepackage{wrapfig}
\usepackage{relsize}
\usepackage{graphicx}
\usepackage{multicol}
\usepackage{multirow}
\usepackage{float}
\usepackage{wrapfig}
\usepackage{multirow}
\usepackage{soul} 
\usepackage{xcolor} 
\definecolor{grayhighlight}{gray}{0.9}

\begin{document}

\title{Breaking Global Self-Attention Bottlenecks in Transformer-based Spiking Neural Networks with Local Structure-Aware Self-Attention}

\author{
Lingdong~Li$^{1}$, Hangming~Zhang$^{2}$, Qiang~Yu$^{2}$\\
{\small $^{1}$School of Future Technology, Tianjin Key Laboratory of Cognitive Computing and Application,}\\
{\small Tianjin University, Tianjin 300354, China}\\
{\small $^{2}$School of Artificial Intelligence, Tianjin Key Laboratory of Cognitive Computing and Application,}\\
{\small College of Intelligence and Computing, Tianjin University, Tianjin 300354, China}\\
{\small Corresponding author: Qiang Yu (e-mail: yuqiang@tju.edu.cn).}
}

\maketitle

\begin{abstract}
Transformer-based Spiking Neural Networks (SNNs) integrate SNNs with global self-attention and have demonstrated impressive performance. However, existing Transformer-based SNNs suffer from two fundamental limitations. First, they typically employ  max pooling layers to reduce the size of feature maps, but the max pooling captures only the strongest response and fails to comprehensively preserve representative regional features. Second, the global self-attention involves all global feature interactions, resulting in computational redundancy and quadratic computational complexity, thus conflicting with the sparse and energy-efficient characteristics of SNNs. To address these challenges, we develop Local Structure-Aware Spiking Transformer (LSFormer), a novel Transformer-based Spiking Neural Network that incorporates Spiking Response Pooling (SPooling) and Local Structure-Aware Spiking Self-Attention (LS-SSA). For the first time, our LSFormer leverages a local dilated window mechanism to capture both local details and long-range dependencies. Experimental results demonstrate that our LSFormer achieves state-of-the-art performance compared to existing advanced Transformer-based SNNs. Notably, on the more challenging static dataset Tiny-ImageNet and neuromorphic dataset N-CALTECH101, LSFormer substantially outperforms state-of-the-art baselines by 4.3\% and 8.6\% in top-1 classification accuracy, respectively. These results highlight the potential of LSFormer to advance energy-efficient spiking models toward practical deployment in large-scale vision applications. 
\end{abstract}


\section{Introduction}
\IEEEPARstart{C}{ompared} with traditional Artificial Neural Networks (ANNs), the brain-inspired Spiking Neural Networks (SNNs) exhibit low power consumption   \cite{Furber2014The,yuan2024trainable,dong2022event,niu2023event,liu2021spiking}, high biological plausibility  \cite{2017A,2014Neuronal,sun2024reliable,tang2025spatio} and strong robustness   \cite{2020LISNN,kim2020spiking,roy2019towards,pei2019towards,zhou2024enhancing,song2021efficient,yu2025}. 
However, the limited performance of SNNs restricts their development and application across various fields, particularly in challenging vision benchmarks.
Researchers have made significant efforts to address the performance bottlenecks of SNNs. The Transformer      \cite{10.5555/3295222.3295349} has demonstrated exceptional performance across diverse domains, including natural language processing \cite{10.5555/3295222.3295349,devlin-etal-2019-bert,gorenstein2024bidirectional,zhang2024survey,buscaldi2024citation,haidarh2025hybrid,MENG2025114270} and computer vision \cite{2021An,dosovitskiy2021image,jiao2023dilateformer,Yuan_2021_ICCV}. Building on the success of Transformer, recent studies have increasingly explored its integration with SNNs to develop Transformer-based SNNs  \cite{fang2024spiking,zhou2023spikformer,zhou2023enhancing,zhou2023spikingformer,shen2024tim,yao2023spikedriven}. These hybrid models combine the powerful ability of Transformer and also leverage the low-energy consumption characteristics of SNNs. Existing Transformer-based SNNs project the original image into spike patch sequence through Spiking Tokenizer (ST) \cite{fang2024spiking,zhou2023spikingformer,shen2024tim}. They then extract spatial and channel features by stacking encoders composed of Spiking Self-Attention (SSA) and MLP block, simultaneously modeling the global dependencies to generate high-quality feature representations. Despite the advances of Transformer-based SNNs over previous SNNs, they still suffer from two critical limitations. 

One limitation is that the Spiking Tokenizer employs max pooling layers to reduce the spatial resolution of feature maps. However, max pooling only retains the strongest response within each pooling window, which may fail to comprehensively capture regional features and potentially result in significant information loss, especially in the presence of sparse spiking activations \cite{sengupta2019going,rathi2020enabling}.
Hybrid pooling \cite{yu2014mixed} and adaptive pooling \cite{lin2013network} have been proposed to combine different feature aggregation strategies. Hybrid pooling simply integrates max pooling and average pooling outputs, but the unfiltered average pooling may weaken sparse peak responses. 
Adaptive pooling relies on global statistics, which can similarly disrupt locally accumulated peak representations.

The other limitation involves the SSA, whose computation depends on dense matrix operations across all global feature interactions.  Previous studies \cite{jiao2023dilateformer,hassani2023neighborhood,liu2021swin}  have shown that task-relevant areas are often locally concentrated, implying that the global receptive field of SSA not only introduces substantial redundant computations but also results in insufficient feature extraction capability, limiting its ability to effectively capture discriminative features. Several studies \cite{hassani2023neighborhood,liu2021swin,liu2022swin} introduce window-based self-attention, restricting attention computation to small neighborhood windows, which effectively reduces irrelevant calculations. However, the performance of window-based self-attention heavily depends on the empirically chosen window size. Specifically, small windows limit the capability to model long-range dependencies, whereas larger windows increase computational and memory costs. Furthermore, the fixed window size lacks flexibility in modeling multiple receptive fields. Consequently, achieving a balance between computational efficiency and performance remains a critical challenge.

In this work, we introduce Spiking Response Pooling (SPooling) to mitigate the significant information loss in Spiking Tokenizer. Unlike max pooling, SPooling effectively exploits the complementary strengths of max pooling and average pooling, enabling the model to capture more comprehensive information within each pooling window. Furthermore, we propose Local Structure-Aware Spiking Self-Attention (LS-SSA) to address the high computational complexity and the limited feature extraction capability of SSA. LS-SSA efficiently captures local details and long-range dependencies across multiple receptive fields without increasing additional computational overhead, and dynamically adjusts channel weights to adaptively emphasize task-relevant channels, improving the quality of feature representation. Based on LS-SSA and SPooling, we develop a Transformer-based SNN, named Local Structure-Aware Spiking Transformer (LSFormer).

It is noteworthy that objects in the real world often exhibit multi-scale structural characteristics, and to recognize such objects accurately requires integrating information across different spatial scales \cite{10131902}. Our LSFormer extracts multi-scale information at different receptive fields, addressing spatial scale variance without increasing computational complexity. The main contributions of this work are summarized as follows:
\begin{itemize}

\item We design a novel Local Structure-Aware Spiking Self-Attention to efficiently capture both fine-grained local details and long-range dependencies without additional computational overhead.

\item We propose Spiking Response Pooling to alleviate the limitations of max pooling and mitigate the loss of effective information during the pooling process, thereby enabling the model to capture more informative features.
\item Extensive experiments demonstrate that our method outperforms state-of-the-art SNNs on various static image datasets, achieving 96.73\% accuracy on CIFAR-10, 82.00\% on CIFAR-100, and 71.61\% on Tiny-ImageNet with only 4 timesteps. On neuromorphic datasets, results on DVS-Gesture (98.6\%), CIFAR10-DVS (84.3\%), and N-CALTECH101 (87.6\%) further validate the generalizability of our method.
\end{itemize}

\section {Preliminaries and Related Works}
\subsection{The Leaky Integrate-and-Fire Neuron}
The Leaky Integrate-and-Fire (LIF) neuron model is widely adopted for its strong biological plausibility and computational efficiency. In this study, we adopt the LIF neuron model as the fundamental unit. The dynamics of a LIF neuron model can be formulated as follows:

\begin{equation}\label{equation22}
v[t]=u[t]+\frac{1}{\tau} \cdot (I[t]-(u[t]-u_\text{reset}))
\end{equation}
\begin{equation}\label{equation23}
s[t]=\Theta \ (v[t]-u_\text{th})
\end{equation}
\begin{equation}\label{equation24}
u[t+1]=s[t]\cdot u_\text{reset}+(1-s[t]) \cdot v[t]
\end{equation}

Here, $t$ denotes the timesteps, and ${\tau}$ denotes the membrane time constant,
which governs the rate of information decay. Eq.~\ref{equation22} describes the charging process, where $u[t]$ and $v[t]$ represent the membrane potential of the postsynaptic neuron before and after charging, respectively. $I[t]$ is the current input, and $u_\text{reset}$ denotes the resting potential.

Eq.~\ref{equation23} describes the firing process, where $\Theta(\cdot)$ is the Heaviside step function, and$\ u_\text{th}$ is the firing threshold. Eq.~\ref{equation24} describes the reset process.

\subsection{Transformer-based Spiking Neural Networks}
Transformer-based SNNs integrate SNNs with Transformer architecture and have attracted significant attention due to their potential for low energy consumption and high performance. Spikformer \cite{zhou2023spikformer} introduces Spiking Self-Attention (SSA), which replaces conventional dense Query, Key, and Value matrices with sparse spike-driven activations. SSA \cite{zhou2023spikformer} effectively eliminates energy-intensive matrix multiplications and the softmax operation, leading to substantial energy savings. Building on this, Spikingformer \cite{zhou2023spikingformer} redesigns the residual connection paradigm, yielding a fully spike-driven architecture that minimizes non-spiking computations. The ConvBN-MaxPooling LIF (CML) downsampling module \cite{zhou2023enhancing} is proposed to address the challenge of insufficient gradient propagation in spiking models. Further advancements include Spike-Driven Transformer \cite{yao2023spikedriven}, which incorporates a novel Spike-Driven Self-Attention (SDSA) mechanism that replaces multiplication operations entirely with masking and addition, thereby reducing computational complexity and energy consumption. QKformer \cite{zhou2024qkformer} simplifies the architecture by retaining only Query and Key matrices, achieving further parameter reduction and energy efficiency while introducing a hierarchical design for the first time in Transformer-based SNNs. Temporal Interaction Module (TIM) \cite{shen2024tim} enhances temporal processing in SNNs by adaptively fusing historical and current information. Additionally, SpikingResformer \cite{shi2024spikingresformer} combines a ResNet-based multi-stage backbone with Dynamic Spiking Self-Attention (DSSA) \cite{shi2024spikingresformer}, improving both accuracy and energy efficiency. SAFormer \cite{zhang2025combining} designs the Spike Aggregated Self-Attention (SASA) mechanism, which significantly simplifies the computation process by calculating attention weights using only the spike matrices query and key, effectively reducing energy consumption.

Despite these advances, existing Transformer-based SNNs predominantly rely on global receptive field self-attention mechanisms, which incur substantial computational redundancy. Moreover, they generally lack explicit mechanisms for modeling multi-scale receptive fields, limiting their effectiveness in complex vision tasks.

\begin{figure*}[t]
  \centering
  \includegraphics[width=0.8\linewidth]{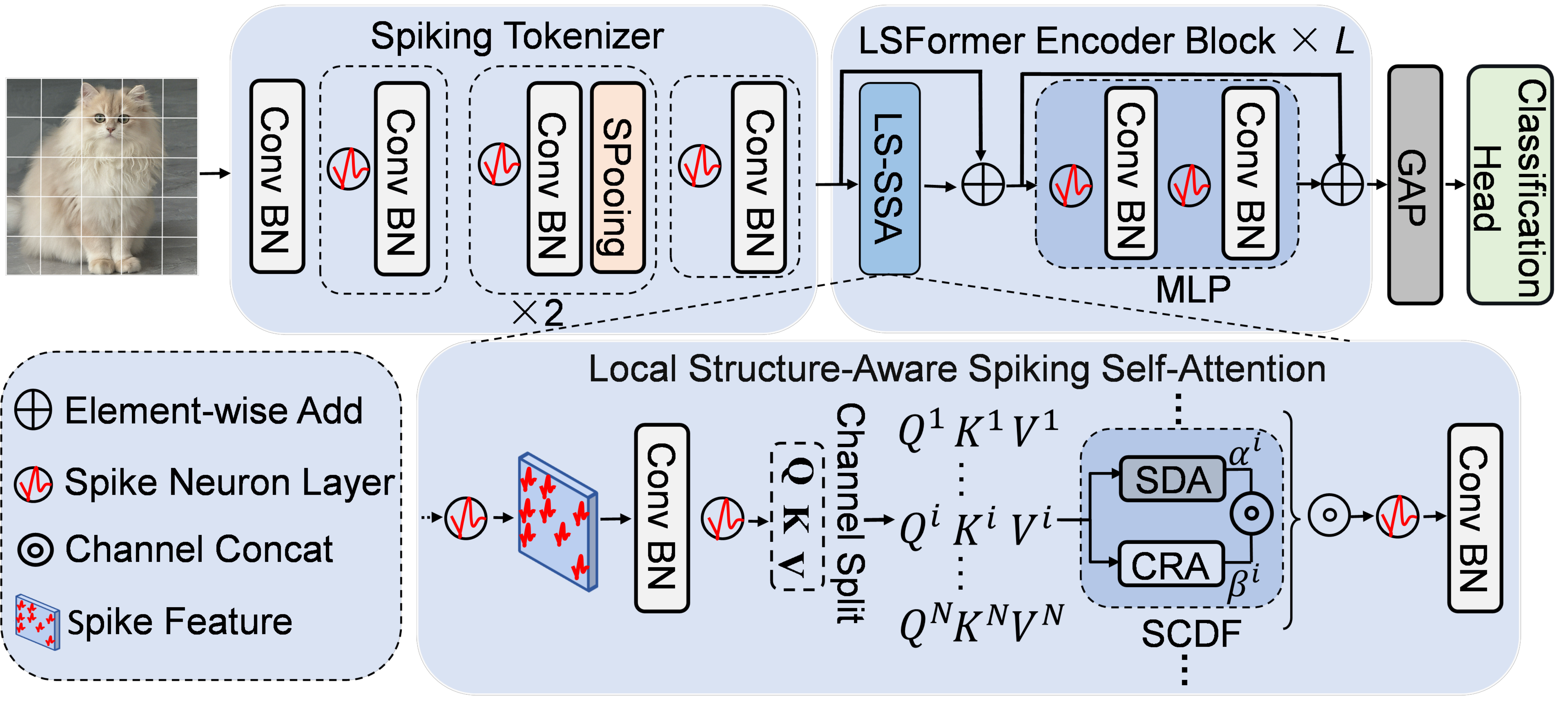}
  \caption{ \label{fig:overview} The overall architecture of our proposed LSFormer. Compared with typical Transformer-based SNNs (e.g., Spikingformer \cite{zhou2023spikingformer}), LSFormer incorporates two key innovations: (1) Local Structure-Aware Spiking Self-Attention, which replaces the conventional SSA to simultaneously capture fine-grained local details and long-range dependencies; (2) Spiking Response Pooling (SPooling) substitutes traditional max pooling to reduce information loss. ConvBN denotes a 2D convolutional layer (stride-1, 3 $\times$ 3 kernel size) followed by batch normalization.  $ \alpha^i$ and $ \beta^i$ denote the learnable parameters of the $i$-th group.}
  
\end{figure*}
\section{Methods}
In this section, we introduce the overall architecture of our proposed LSFormer (\mbox{Section~\ref{overview}}), the novel downsampling mechanism Spiking Response Pooling (\mbox{Section~\ref{Spiking Response Pooling}}) and the Local Structure-Aware Spiking Self-Attention (\mbox{Section~\ref{Local Structure-Aware Spiking Self-Attention}}).

\subsection{Overall Architecture of LSFormer}  \label{overview}
As illustrated in Fig.~\ref{fig:overview}, our proposed LSFormer contains a Spiking Tokenizer (ST), a stack of $L$ encoder blocks, and a linear classification head. Given a 2D image sequence $I\in  \mathbb{R}^{T\times C\times H\times W}$, where $T$, $C$, $H$, and $W$ denote timesteps (for static image datasets, $T$ is the number of repetitions), channel, height and width of image sequence, respectively. We employ the Spiking Tokenizer to project the image sequence into the $D$-dimensional spike patch sequence $U_{0}$. 
\begin{equation}\label{equation25}
U_{0}=\operatorname{ST}(I),\ I\in \mathbb{R}^{ T\times C\times H\times W},\ U_{0}\in \mathbb{R}^{ T\times D\times H^{'}\times W^{'}}
\end{equation}
Here, $H^{'}$ and $W^{'}$ represent the height and width of $U_{0}$, respectively.
For static image datasets, the values of $H^{'}$ and $W^{'}$ are obtained through two SPooling operations (stride-2, 3 $\times$ 3 kernel size) in the Spiking Tokenizer. The spatial dimension of $U_0$ after the SPooling operations can be calculated with reference to Eq.~\ref{equation3}. $D$ represents the channel dimension of $U_{0}$ (according to \cite{zhou2023spikformer,shi2024spikingresformer,zhou2024spikformer,shen2024tim,fang2023spikingjellyopensourcemachinelearning,zhou2023spikingformer,fang2024spiking}, $D$ is set to 384 for static datasets and 256 for neuromorphic datasets). Specifically, the Spiking Tokenizer incorporates four convolutional layers, whose output channels are $D/8$, $D/4$, $D/2$, and $D$ respectively, gradually expanding to the channel dimension $D$ of $U_{0}$.
Then we pass $U_{0}$ to a stack of $L$ encoder blocks. Each encoder block consists of a LS-SSA (Section~\ref{Local Structure-Aware Spiking Self-Attention}) and a MLP block \cite{zhou2023spikingformer}. Residual connection is employed for membrane potential in both the LS-SSA and the MLP. Signal flow is formulated as:

\begin{equation}\label{equation28}
U_{l} = \operatorname{LS-SSA}(U_{l-1}) + U_{l-1}, U_{l} \in \mathbb{R}^{T\times D\times H^{'}\times W^{'}}, l=1,\cdots,L 
\end{equation}

\begin{equation}\label{equation29}
S_{l} = \operatorname{MLP}(U_{l}) + U_{l}, S_{l} \in \mathbb{R}^{ T\times D\times H^{'}\times W^{'}}, l=1,\cdots,L 
\end{equation}
Here, $U_{l}$ and $S_{l}$ represent the outputs of the $l$ th layer of the LS-SSA and the encoder block, respectively. Finally, a global average-pooling (GAP) is utilized on the output of the last encoder block to yield $D$-dimensional feature, which is sent to the fully-connected-layer classification head (CH) to output the prediction $Y$.
\begin{equation}\label{equation30}
Y = \operatorname{CH}(\operatorname{GAP}(S_{L}))
\end{equation}

\subsection{Spiking Response Pooling}\label{Spiking Response Pooling}
In SNNs, spike responses are inherently sparse and locally distributed, and their discriminative information often lies in sub-threshold activations, which do not necessarily trigger spikes. These sub-threshold activations reflect accumulated neuronal states and thus constitute a complementary source of task-relevant information \cite{hwang2020impact}.
Conventional max pooling only retains the peak response within each pooling window,  potentially discarding weak but informative spikes that are critical for effective information integration. 
This issue is particularly severe in directly trained SNNs, where spike firing rates are typically low. Motivated by this, we design SPooling to explicitly preserve representative spike responses by combining peak spikes and  informative sub-threshold activations.

Given the input sequence $x\in \mathbb{R} ^{T\times C\times H\times W}$, where $T$, $C$, $H$, and $W$ denote timesteps, channel, height and width of input sequence $x$, respectively.
The max pooling operation selects the maximum value within a defined pooling window $ R_{i,j}$ centered at position $(i,j)$. The process is formulated as:
\begin{equation}\label{equation1}
x_{\text{max}}(i,j)=\underset{(a,b)\in R_{i,j} }{\operatorname{Max}} x(a,b) 
\end{equation}

The average pooling operation calculates the average response within the same pooling window:
\begin{equation}\label{equation2}
x_{\text{avg}}(i,j)= \frac{1}{k\times k }\sum_{(a,b)\in R_{i,j}}x(a,b) 
\end{equation}

Here, $k\times k$ defines the size of the pooling window. $x(a,b)$ represents the feature value at coordinate $(a,b)$ in $R_{(i,j)}$. $x_{\text{max}}(i,j)$ and $x_{\text{avg}}$ (both $\in  \mathbb{R} ^{T\times C\times H_{1}\times W_{1}}$) represent the result of max pooling and average pooling, respectively. After the pooling operation, the height ($H_{1}$) and width ($W_{1}$) of the feature map can be determined as Eq.~\ref{equation3}. 
\begin{equation}\label{equation3}
\small
H_{1} =\left \lfloor \frac{H+2p-k}{s}\right \rfloor +1  \ \ \  W_{1} =\left \lfloor \frac{W+2p-k}{s} \right \rfloor +1
\end{equation}

\begin{figure}[!t]
\centering
\includegraphics[width=0.9\linewidth]{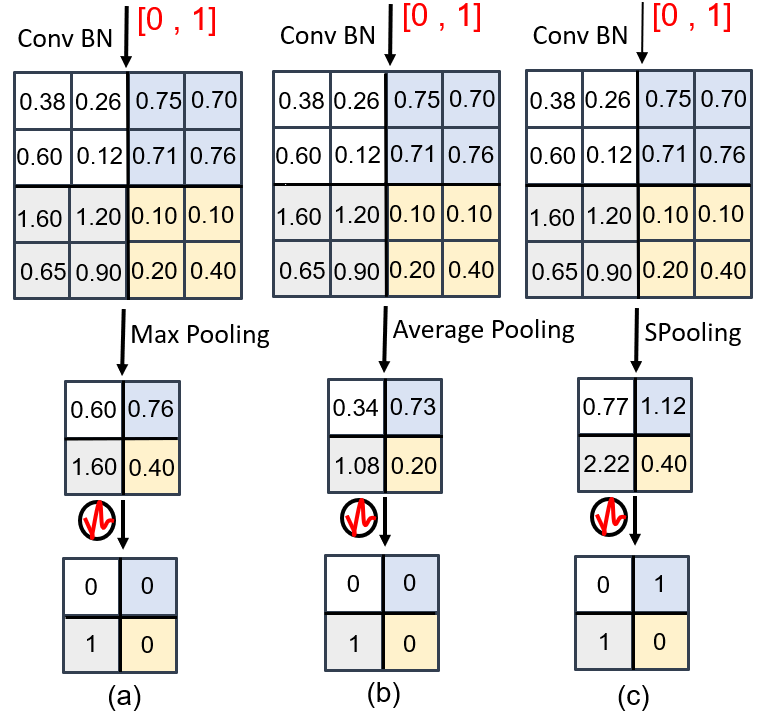}
\caption{\label{fig:pooling} The downsampling demonstration of max pooling, average pooling, and our proposed SPooling. (a) shows the max pooling results within the pooling windows. (b) illustrates the average pooling results within the same windows. (c) shows the SPooling results, which integrates both the maximum and average responses. Specifically, blue and yellow region: when $x_{\text{avg}} > \lambda$, $x_{\text{avg}}' = x_{\text{avg}}$,  $x = x_{\text{max}} + \sigma(\theta)\cdot x_{\text{avg}}'$. Green and white region: when $x_{\text{avg}} < \lambda$, $x_{\text{avg}}' = 0$, $x = x_{\text{max}}$.}
\end{figure}

Here, $p$ denotes padding, and $s$ represents the stride of the pooling operation.

To effectively identify and preserve informative sub-threshold responses, our SPooling introduces a threshold $\lambda$ to evaluate the average response level within each pooling window. This evaluation process is formulated as: 
\begin{equation}\label{equation4}
    x_{\text{avg}}' (i,j) = 
    \begin{cases}
        x_{\text{avg}}(i,j), &  \ \ \ x_{\text{avg}}(i,j) \geq \lambda \\
        \ \ \ \ \ \ \ 0, &  \ \ \ x_{\text{avg}}(i,j) \textless \lambda
    \end{cases}
\end{equation}

Here, $x_{\text{avg}}$ represents the average pooling result, and $x_{\text{avg}}'$ represents the filtered average response. As shown in Eq.~\ref{equation4}, when  $x_{\text{avg}}$ exceeds the threshold $\lambda$, it is considered to contain meaningful features and thus retained. In contrast, the values of $x_{\text{avg}}$ below the threshold are filtered out.

Furthermore, we introduce a learnable parameter $\theta$ to dynamically regulate the contribution of $ x_{\text{avg}}'$. The final output of SPooling is computed as Eq.~\ref{equation5}.

\begin{equation}\label{equation5}
x(i,j)=
x_{\text{max}}(i,j)+\sigma (\theta )\cdot  x_{\text{avg}}'(i,j) 
\end{equation}

Here, $\theta$ is the learnable gating parameter, and $\sigma (\cdot)$ is the sigmoid function, which constrains $\theta$ within the range $(0,1)$ to ensure stable gradient propagation. The resulting differences among SPooling, max pooling, and average pooling are illustrated in Fig.~\ref{fig:pooling}.

\subsection{Local Structure-Aware Spiking Self-Attention} \label{Local Structure-Aware Spiking Self-Attention}
The Local Structure-Aware Spiking Self-Attention integrates the Spatial Dilated Attention (SDA), the Channel Recalibration Attention (CRA), and the Spatial-Channel Decoupled Fusion Architecture. Given the spike-form sequence $X$ $\in \{0,1\}^{T\times D\times H^{'}\times W^{'}}$,  where $T$, $D$, $H^{'}$, and $W^{'}$ represent the timesteps, channel, height and width of the $X$, respectively. $X$ is first mapped to generate spike-form query (Q), key (K), and value (V) matrices. This mapping process is formulated as:

\begin{equation}\label{equation6}
Q = \operatorname{SN}_{Q}\left(\operatorname{BN}_{Q}(\operatorname{Conv}_{Q}(X))\right)
\end{equation}
\begin{equation}\label{equation7}
K = \operatorname{SN}_{K}\left(\operatorname{BN}_{K}(\operatorname{Conv}_{K}(X))\right)
\end{equation}
\begin{equation}\label{equation8}
V = \operatorname{SN}_{V}\left(\operatorname{BN}_{V}(\operatorname{Conv}_{V}(X))\right)
\end{equation}

Here, $Q$, $K$, and $V$ $\in \{0,1\}^{T\times D\times H^{'}\times W^{'}}$, retaining the same dimensions as the input sequence $X$. $\textup{BN}$ and $\textup{Conv}$ represent batch normalization and two-dimensional convolution operations, respectively. $\text{SN}(\cdot)$ means the spiking neuron layer. 

Given that the $Q$, $K$, and $V$ matrices are uniformly partitioned into $N$ groups along the channel dimension. For the $m$-th group $(m\in \{1,2,\cdots,N\})$, the query, key, and value matrix slices are defined as:
\begin{equation}\label{equation9}
Q^{m}=\Gamma (Q,m),\ \ \ K^{m}=\Gamma(K,m),\ \ \ V^{m}=\Gamma(V,m)
\end{equation}

Here, $\Gamma(M,m)$ denotes the slicing operation that extracts the $m$-th group from matrix $M$ along channel dimension, where $M$ $\in \{Q,K,V\}$. $Q^{m}$, $K^{m}$, and $V^{m}$ represent the $m$-th group slices of the original $Q$, $K$, $V$ matrices, respectively. These slices preserve the spatial dimensions ($H^{'}$ and $W^{'}$) and timesteps ($T$) of their corresponding original matrices, while their channel dimension is reduced to $d = D/N$ (where $D$ represents the channel dimension of each original $Q$, $K$, and $V$ matrix).

\subsection{Spatial Dilated Attention}
SNNs process information through discrete spike events and rely on the temporal accumulation of membrane potentials, resulting in responses that are sparse and highly localized in spatial dimension. Due to this, applying global attention in SNNs results in ineffective spike interactions, which not only increase computational redundancy but also weaken meaningful spike accumulation. Unlike traditional SSA mechanisms \cite{shi2024spikingresformer,yao2023spikedriven,zhou2023spikformer,zhou2024spikformer}, which rely on dense feature interactions, our SDA is specifically designed to accommodate spike-driven computation. By restricting attention computation to local receptive fields, SDA enhances the utilization efficiency of sparse spike responses while still preserving long-range dependencies through multi-scale dilated windows.

\begin{figure*}[t]
\centering
\includegraphics[width=0.9\linewidth]{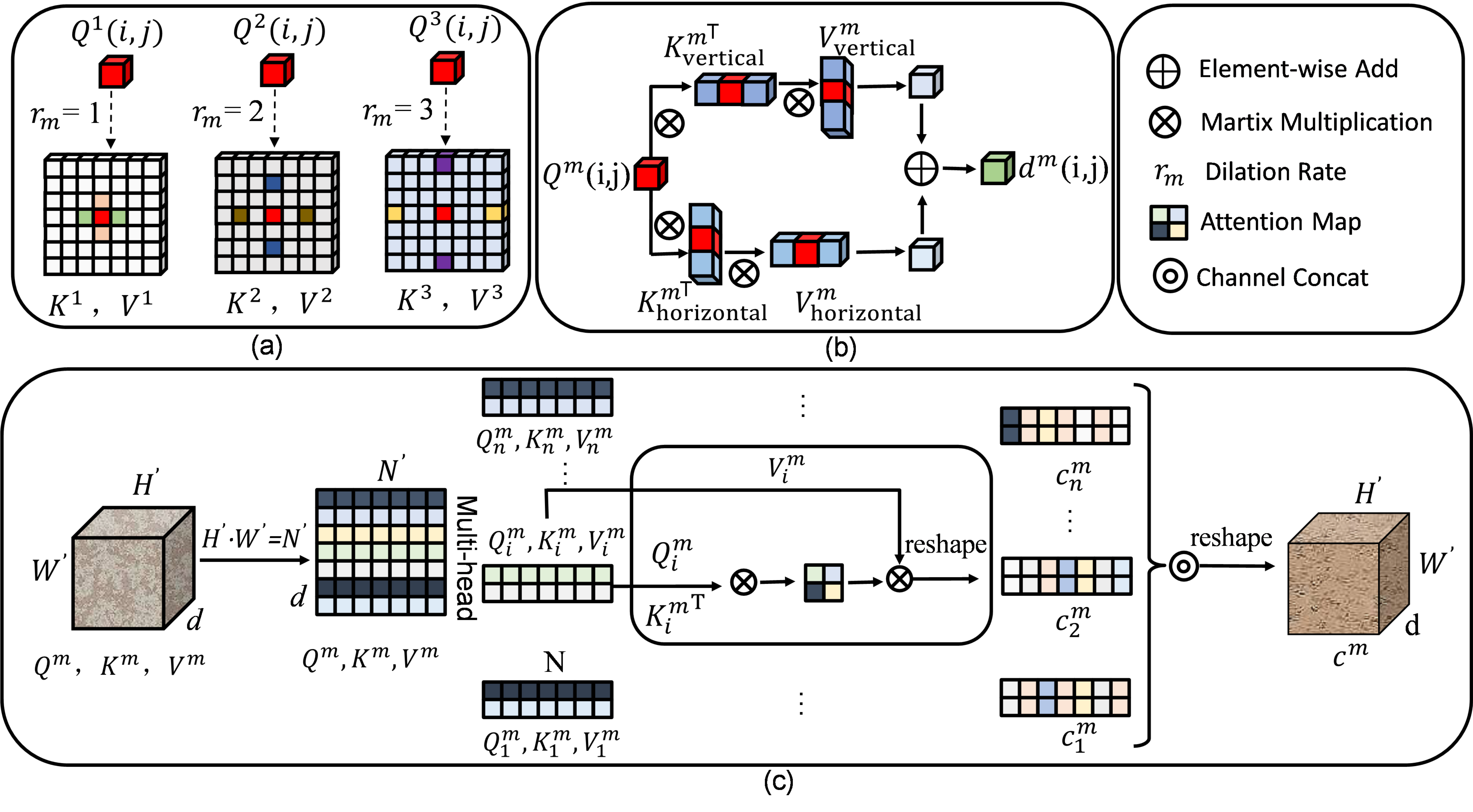}
\caption{\label{fig:attn} The detailed implementations of SDA and CRA. (a) illustrates that the query matrix slices sparsely select key and value matrices in the horizontal and vertical windows via different dilation rates $r_m$ (1, 2, and 3). (b) shows the computation process of attention in both horizontal and vertical windows within the $m$-th group. (c) shows the computation process of Channel Recalibration Attention in the $m$-th group.}
\end{figure*}

As illustrated in Fig.~\ref{fig:attn} (a), SDA assigns distinct dilation rates ($r_m$) to different groups, enabling sparse sampling of the corresponding  $K^{m}$ and $V^{m}$ slices from receptive fields of varying scales along the horizontal and vertical directions, thereby reducing computational overhead while enabling more effective capture of directional structural patterns in spatial features. This design allows small receptive fields (with smaller $r_m$) to sparsely sample local spatial regions and capture fine-grained features, while large receptive fields (with larger $r_m$) cover wider spatial ranges and model long-range dependencies.

For the query matrix slice $Q^m$ at spatial position $(i,j)$, the sampling points of corresponding $K^m$ and $V^m$ within the horizontal window are defined as  Eq.~\ref{equation11} and Eq.~\ref{equation11_2}. 
\begin{equation}\label{equation11}
K_{\text{horizontal}}^{m} = \left[ K^{m}(i, j + \delta_\text{h}\cdot r_m) \mid \delta_\text{h} \in \mathcal{R}_\text{h} \right]
\end{equation}
\begin{equation}\label{equation11_2}
V_{\text{horizontal}}^{m} = \left[ V^{m}(i, j+ \delta_\text{h}\cdot r_m) \mid \delta_\text{h} \in \mathcal{R}_\text{h} \right]
\end{equation}

Constructing the vertical window to sample $K^m$ and $V^m$ along the vertical direction, the sampling points are defined as Eq.~\ref{equation10} and Eq.~\ref{equation10_2}.
\begin{equation}\label{equation10}
K_{\text{vertical}}^{m} = \left[ K^{m}(i + \delta_\text{v} \cdot r_m, j) \mid \delta_\text{v} \in \mathcal{R}_\text{v} \right]
\end{equation}
\begin{equation}\label{equation10_2}
V_{\text{vertical}}^{m} = \left[ V^{m}(i + \delta_\text{v}\cdot r_m, j) \mid \delta_\text{v} \in \mathcal{R}_\text{v} \right]
\end{equation}

Here, $\delta_\text{h}$ and $\delta_\text{v}$ denote the offset parameters for sampling positions in the horizontal and vertical windows relative to $Q^m$, respectively. $r_m$ represents the dilation rate. $\mathcal{R}_\text{h} $ and $\mathcal{R}_\text{v}$ define the sampling ranges in the horizontal and vertical windows, as described in Eq.~\ref{equation12} and~\ref{equation12_2}.
\begin{equation}\label{equation12}
 \mathcal{R}_\text{h}=\left \{ - \left \lfloor \frac{w_{\text{h}} }{2} \right \rfloor  , - \left \lfloor \frac{w_{\text{h}} }{2} \right \rfloor +1 ,\cdots, \left \lfloor\frac{w_{\text{h}}}{2} \right \rfloor \right \} 
 \end{equation}
 \begin{equation}\label{equation12_2}
 \mathcal{R}_\text{v} =\left \{ - \left \lfloor \frac{w_{\text{v}} }{2} \right \rfloor  ,- \left \lfloor \frac{w_{\text{v}} }{2} \right \rfloor+1,\cdots, \left \lfloor\frac{w_{\text{v}}}{2} \right \rfloor \right \} 
\end{equation} 

Here, $\left \lfloor \cdot \right \rfloor$ denotes the floor operation. $w_\text{h}$ and $w_\text{v}$ are the sizes of the horizontal and vertical sliding windows, respectively. For instance, when the $w_\text{h}$ and $w_\text{v}$ are 3, both $\mathcal{R}_\text{h}$ and $\mathcal{R}_\text{v}$ are $\left \{ -1,0,1\right\}$.

Notably, SDA implements a multi-scale attention mechanism by assigning a unique dilation rate $r_m$ to each group obtained through the partition strategy defined in Eq.~\ref{equation9}, where $Q$, $K$, and $V$ are uniformly partitioned into $N$ groups along the channel dimension. When $N=1$, SDA degenerates to a single-scale attention mechanism. Specifically, for different groups, distinct dilation rates are employed to adjust the spatial spacing of the sampled \(K^m\) and \(V^m\) points, thereby enabling sampling from receptive fields of varying scales. In our experiments, $Q$, $K$, and $V$ matrices are divided into three groups with dilation rates of 1, 2, and 3, respectively. 

As shown in Fig.~\ref{fig:attn} (b), the self-attention operations are independently computed within the horizontal and vertical windows. The process is described as Eq.~\ref{equation13}. 
\begin{equation}\label{equation13}
\small
\begin{aligned}
d^m =  \left(Q^{m}(K_{\text{horizontal}}^{m})^{\text{T}} V_{\text{horizontal}}^{m} \ast s^{\ast} \right) \oplus \ \left(Q^{m} (K_{\text{vertical}}^{m})^{\text{T}} V_{\text{vertical}}^{m} \ast s^{\ast}\right)
\end{aligned}
\end{equation}

Here, $d^{m}$ represents the aggregate features of the horizontal and vertical directions in the $m$-th group. $\oplus$ represents element addition operation.  $s^{\ast}$ serves as scaling factor to constrain the large value of the matrix multiplication result, calculated as \(1/\sqrt{d}\),  a standard approach consistent with that proposed in \cite{10.5555/3295222.3295349}, where $d$ represents the channel dimension of $Q^{m}$.

During the sampling of boundary query slices, we adopt a zero-padding strategy to these slices as proposed in \cite{jiao2023dilateformer}. However, this approach leads to the incomplete feature extraction \cite{jiao2023dilateformer}. To address this issue, we apply depthwise separable convolution (DWC) \cite{hu2024advancing, hao2023progressive} to the value matrix slice $V^m$ for complementing the edge region features, as formally described in Eq.~\ref{eq:dwc}.

\begin{equation}\label{eq:dwc}
\text{DWC}(V^m) = \text{BN}\left( V^m _{\ast_{\text{dw}}} \mathbf{W}_{\text{dw}} \right)
\end{equation}

Here, $\text{DWC}(\cdot)$ represents the depthwise separable convolution operation. $ V^m_{\ast_{\text{dw}}} \mathbf{W}_{\text{dw}}$ refers to the depthwise convolution of $V^m$ with the weight matrix ${W}_{\text{dw}}$.
The final output of SDA, denoted as $s^{m}$, is defined in Eq.~\ref{equation14}.
\begin{equation}\label{equation14}
s^{m} = d^{m} + \operatorname{DWC}(V^{m}), s^m\in \mathbb{R}^{T\times d\times H^{'}\times W^{'}}
\end{equation}

\subsection{Channel Recalibration Attention}
Naive channel-wise modulation may weaken representation of important features across channels, which fails to ensure uniform activation of task-relevant channels.  For example, in a feature map where only edge regions of an object emit spikes, SSA assigns significant attention to inactive background channels, reducing feature representation quality.  To address this issue, we propose CRA, which effectively enhances the representation of task-relevant channel-wise features. As shown in Fig.~\ref{fig:attn} (c), given $Q^{m}$, $K^{m}$, and $V^{m}$ in the $m$-th group, we first flatten their spatial dimensions ($H^{'}$ and $W^{'}$) into one dimension $N'$ (\(N' = H^{'} \times W^{'}\)), then split them into $n$ heads along the channel dimension to reduce the computational complexity \cite{zhou2023spikformer,zhou2023spikingformer,shen2024tim}.

For the $m$-th group and the $h$-th head, the \(Q_h^m\), \(K_h^m\), and \(V_h^m\) slices are represented as:

\begin{equation}\label{equation15}
Q_h^m, K_h^m, V_h^m \in \mathbb{R}^{T\times N'\times \frac{d}{n}}
\end{equation}

Here, $T$, $N'$ and $d/n$ denote the timesteps, spatial dimension, and channel dimension of \(Q_h^m\), \(K_h^m\), and \(V_h^m\), respectively. Specifically, $d/n$ is obtained by dividing $d$ into $n$ equal parts ($n$ is the number of attention heads).  The attention calculation process for the $m$-th group and the $h$-th head is described as Eq.~\ref{equation16}. Meanwhile, this same procedure applies to all other heads.
\begin{equation}\label{equation16}
c_h^{m} =Q _h^{m}  (K _h^{m} )^ \textup{T} V _h^{m}\ast s^{\ast}
\end{equation}

Here, $c_h^{m}$ represents the attention output of the $h$-th head in the $m$-th group. $s^{\ast}$ denotes the scaling factor, same as in Eq.~\ref{equation13}. The outputs of all heads are then concatenated along the channel dimension and  the flattened spatial dimension \(N'\) is reshaped back to the original $H^{'}$ and $W^{'}$ to form the final result $c^{m}$ of CRA in the $m$-th group.
\begin{equation}\label{equation17}
c^{m} = \operatorname{concat}\left \{ c_{{1}}^{m}, c_{{2}}^{m}, \cdots , c_{n}^{m} \right \} 
\end{equation}

Here, $c^{m}\in \mathbb{R} ^{T\times d\times H^{'}\times W^{'}}$ represents the final output of CRA, where $T$, $d$, $H^{'}$, and $W^{'}$ denote timesteps, channel, height and width of $c^{m}$. $\operatorname{concat} \left \{ c_{h}^{m}  \right \} _{h=1}^{n}$  represents concatenation across all heads within the $m$-th group.

\subsection{Spatial-Channel Decoupled Fusion Architecture}\label{3.5}
Existing Spiking Self-Attention mechanisms \cite{yao2023spikedriven,zhou2023spikformer,zhou2024spikformer} often struggle to simultaneously capture spatial dependencies and channel-wise discriminative features. We propose SCDF to effectively exploit both spatial and channel-wise features. As illustrated in Fig.~\ref{fig:overview}, SCDF consists of two parallel branches, the SDA branch and the CRA branch, which extract complementary information from spatial and channel dimensions, respectively. The contributions of the two branches are adaptively balanced through learnable parameters, allowing the model to dynamically balance spatial and channel-wise features adaptively, emphasizing the critical information from their respective dimensions.

Specifically, for the SDA branch, each group is assigned a unique dilation rate \(r_m\), enabling sparse sampling of the feature map with different receptive fields, achieving scale-specific spatial feature extraction, while CRA focuses on channel-wise feature recalibration to enhance the representation of task-relevant channels.
\begin{equation}\label{equation18}
A ^{m} =\operatorname{concat} \left \{ \alpha ^{m} s^{m} \ , \  \beta  ^{m} c^{m} \right \} 
\end{equation}

Here, $\alpha^{m}$ and $\beta^{m}$ in Eq.~\ref
{equation18} are learnable parameters in the $m$-th group. To avoid introducing prior bias toward either spatial aggregation or channel-wise enhancement, all $\alpha^{m}$ and $\beta^{m}$ are initialized to $0.5$, ensuring balanced contributions from both branches at the early stage of training.
These parameters are then optimized jointly with the rest of the network through backpropagation. \(A^{m} \in \mathbb{R}^{T\times 2d\times H^{'}\times W^{'}}\) denotes the spatial-channel decoupled features for the $m$-th group. The size of $A^{m}$ is determined by \(s^m\) and \(c^m\), both of which have a size of \(\mathbb{R}^{T\times d\times H^{'}\times W^{'}}\). After concatenation along the channel dimension, the channel dimension doubles from $d$ to $2d$, while  $T$, \(H^{'}\), and \(W^{'}\) remain unchanged. Then, the decoupled features from $N$ independent groups are concatenated along the channel dimension to aggregate multi-scale spatial-channel
features.
\begin{equation}\label{equation19}
Att =\operatorname{concat} \left \{ A ^{1} ,A ^{2}, \cdots , A ^{N}  \right \}
\end{equation}

Here, $ \operatorname{concat} \left \{ A^{m}  \right \} _{m=1}^{N}$ denotes channel-wise concatenation from $N$ groups. The result $Att$ is processed through the spike neuron and the convolution layer to aggregate the features. The final result of the LS-SSA can be described as Eq.~\ref{equation20}. 
\begin{equation}\label{equation20}
x' = \operatorname{BN}(\operatorname{Conv}(\operatorname{SN}(Att)))
\end{equation}

Here, $x'$ represents the final output in LS-SSA. $ \operatorname{SN}\left ( \cdot  \right ) $ represents the spike neuron layer. $ \operatorname{Conv}\left ( \cdot  \right ) $ denotes the two-dimensional convolution, and $\operatorname{BN}\left ( \cdot  \right ) $ performs batch normalization to stabilize feature distributions.

\section{Experiments}
In this section, we first introduce the theoretical calculation methods for synaptic operation count and energy consumption (Section~\ref{Theoretical}). To comprehensively evaluate the performance and generalization capability of our LSFormer, we conduct experiments on both static image datasets and neuromorphic datasets. Static image datasets, including CIFAR-10 \cite{alex2009learning}, CIFAR-100 \cite{alex2009learning}, and Tiny-ImageNet \cite{alex2009learning} (Section~\ref{4.1}), serve as standard benchmarks for visual recognition tasks, encompassing a range of complexity and scale. These datasets assess the ability of our LSFormer to support accurate classification in conventional vision scenarios. In contrast, the neuromorphic datasets, including CIFAR10-DVS \cite{li2017cifar10}, DVS128 Gesture \cite{amir2017low}, and N-CALTECH101 \cite{orchard2015converting} (Section~\ref{4.2}), provide event-based and sparse spiking inputs. Evaluating LSFormer on these datasets demonstrates its capacity to effectively leverage temporal spike information and process event-driven data. The datasets and implementation details are provided in the Appendix~\ref{Dataset}. Additionally, we analyze the computational complexity of the LS-SSA in the Appendix~\ref{Computational}. Our attention visualization analysis of LSFormer proves its superiority (Section~\ref{4.4}). Furthermore, we conduct ablation studies to assess the robustness and generalizability of the LS-SSA and SPooling (Section~\ref{4.3}). We perform three experimental trials and compute the mean values to avoid randomness. Our experimental results are presented as Top-1
accuracy (\%) with mean ± standard deviation values.

\begin{table*}[t]
\centering
\caption{Experimental results on CIFAR-10 and CIFAR-100. Note that LSFormer-4-384 means LSFormer contains four encoder blocks and 384 feature embedding, `T' denotes `Timesteps', `Accuracy' denotes `Top-1 Classification Accuracy (\%)', `Param' refers to `Parameters (M)'.}
\label{tab:performance1}
\setlength{\tabcolsep}{12pt}

\begin{tabular}{l l c c c c}
\toprule
\multirow{2}{*}{\textbf{Method}} 
& \multirow{2}{*}{\textbf{Architecture}} 
& \multirow{2}{*}{\textbf{Param}} 
& \multirow{2}{*}{\textbf{T}} 
& \multicolumn{2}{c}{\textbf{Accuracy}} \\
\cmidrule(r){5-6}
& & & & \textbf{CIFAR-10} & \textbf{CIFAR-100} \\
\midrule
SALT \cite{kim2021optimizing}                & VGG-9/11              & --    & 40 & 87.10 & 59.11 \\
STDP \cite{wu2018spatio}                    & CIFARNet              & 17.54 & 12 & 89.83 & --    \\
STBP-tdBN \cite{zheng2021going}             & ResNet-19             & 12.63 & 4  & 92.92 & 70.86 \\
ResNet \cite{SHEN2023119170}                & ResNet-19             & 12.63 & 1  & 94.97 & 75.35 \\
SEW-ResNet \cite{fang2021deep}              & SEW-ResNet-21/18      & --    & 16 & 90.02 & 77.37 \\
DSR \cite{meng2022training}                 & PreAct-ResNet-18      & --    & 20 & 95.40 & 78.50 \\
Transformer \cite{dosovitskiy2021image}  & Transformer-4-384     & 9.32  & 1  & 96.73 & 81.02 \\
TET \cite{deng2022temporal}                 & ResNet-19             & 12.63 & 4  & 94.44 & 74.47 \\
Spikformer \cite{zhou2023spikformer}        & Spikformer-4-384      & 9.32  & 4  & 95.51 & 78.21 \\
Spikingformer \cite{zhou2023spikingformer}  & Spikingformer-4-384   & 9.32  & 4  & 95.81 & 79.21 \\
S-Transformer \cite{yao2023spikedriven}     & Spikformer-4-384      & 10.28 & 4  & 95.60 & 78.40 \\
Spikformer (CML) \cite{zhou2023enhancing}   & Spikformer-4-384      & 9.32  & 4  & 96.04 & 80.02 \\
Spikingformer (CML) \cite{zhou2023enhancing}& Spikingformer-4-384   & 9.32  & 4  & 95.95 & 80.37 \\
QKformer \cite{zhou2024qkformer}             & HST-4-384             & --    & 4  & 96.18 & 81.15 \\
SWformer \cite{fang2024spiking}              & SWformer-4-384        & --    & 4  & 96.10 & 79.30 \\
SAFormer \cite{zhang2025combining}           & SAFormer-4-384        & 9.32  & 4  & 95.80 & 79.07 \\
\midrule
\textbf{LSFormer (Ours)}                              & \textbf{LSFormer-4-384}        & \textbf{9.50}  & \textbf{4}  & \textbf{$\textbf{96.73} \pm \textbf{0.07}$} & \textbf{$\textbf{82.00} \pm \textbf{0.04}$} \\
\bottomrule
\end{tabular}
\end{table*}

\subsection{Theoretical Calculations of Synaptic Operation Count and Energy Consumption} \label{Theoretical}
To evaluate the theoretical energy consumption of LSFormer, we first calculate the number of synaptic operations (SOPs):
\begin{equation}\label{sops_calculation}
\text{SOPs}(l) = fr \times T \times \text{FLOPs}(l)
\end{equation}

Here, $l$ is a block in LSFormer, \(fr\) is the firing rate of the input spike sequence, and $T$ is the spike neuron simulation timesteps. \(\text{SOPs}(l)\) refers to the number of synaptic operations at block $l$. \(\text{FLOPs}(l)\) refers to the floating-point operations of block $l$. Energy consumption is proportional to the number of operations. The theoretical energy consumption
of LSFormer is calculated as:
\begin{equation}\label{sops_calculation}
\text{Power} (l) = 77 \text{\ fJ} \times \text{SOPs}(l)
\end{equation}where, 77 fJ (femtojoule) denotes the energy cost per synaptic operation \cite{hu2021spiking}.

For ANNs, the energy cost is directly related to the number of FLOPs, calculated as:
\begin{equation}\label{sops_calculation-ann}
\text{Power} (l) = 12.5 \text{\ pJ} \times \text{FLOPs}(l)
\end{equation} where 12.5 pJ (picojoule) represents the energy consumption per FLOP. The theoretical energy consumption analysis of our LSFormer and the compared methods is shown in the Appendix~\ref{SOPs}.

\subsection{Results and Analysis on Static Image Datasets}\label{4.1}
As shown in Table~\ref{tab:performance1}, LSFormer outperforms all the models of Transformer-based SNNs with the same number of parameters. On the CIFAR-10 dataset, LSFormer achieves a top-1 classification accuracy of 96.73\% with 4 timesteps, significantly surpassing Spikformer \cite{zhou2023spikformer} and Spikingformer \cite{zhou2023spikingformer} by 1.22\% and 0.92\%, respectively.  Additionally, compared with QKformer \cite{zhou2024qkformer}, LSFormer achieves a top-1 classification accuracy improvement of 0.55\%. On the CIFAR-100 dataset, LSFormer attains a top-1 classification accuracy of 82.00\%, outperforming Spikingformer \cite{zhou2023spikingformer} and QKformer \cite{zhou2024qkformer} by 2.79\% and 0.85\%, respectively. To the best of
our knowledge, our LSFormer achieves the state-of-the-art in directly trained pure spike-driven SNN model on both CIFAR10 and CIFAR100 datasets.
\begin{table}[h]
\centering
\small
\caption{Experimental results on Tiny-ImageNet. Note that `T' denotes `Timesteps', `Accuracy' denotes `Top-1 Classification Accuracy (\%)'.}
\label{tab:performance_tiny_imagenet}

\begin{tabular}{lccc}
\toprule
\multirow{2}{*}{\textbf{Method}} & \multirow{2}{*}{\textbf{Architecture}} & \multicolumn{2}{c}{\textbf{Tiny-ImageNet}} \\
\cmidrule(r){3-4}
& & \textbf{T} & \textbf{Accuracy} \\
\midrule
ASGL \cite{wang2023adaptive} & VGG-13 & 4 & 56.57 \\
DCT \cite{garg2021dct} & VGG-13 & 125 & 56.90 \\
Online-LTL \cite{yang2022training} & VGG-13 & 16 & 54.82 \\
Offline-LTL \cite{yang2022training} & VGG-13 & 16 & 55.37 \\
LM-H \cite{hao2023progressive} & VGG-13 & 4 & 59.51 \\
LM-HT(L=2) \cite{hao2024lm} & VGG-13 & 4 &  61.75 \\
LM-HT(L=4) \cite{hao2024lm} & VGG-13 & 2 & 62.05 \\
Spikformer \cite{zhou2023spikformer} & Spikformer-4-384 & 4 & 66.03 \\
S-Transformer \cite{yao2023spikedriven} & Spikingformer-4-384 & 4 & 65.71 \\
SAFormer \cite{zhang2025combining} & SAFormer-4-384 & 4 & 67.28 \\
\midrule
\textbf{LSFormer (Ours)} & \textbf{LSFormer-4-384} & \textbf{4} & \textbf{$\textbf{71.61} \pm \textbf{0.21}$} \\
\bottomrule
\end{tabular}
\end{table}

To further investigate the generalization capability of our LSFormer, we evaluate its performance on the more challenging Tiny-ImageNet dataset. As shown in Table~\ref{tab:performance_tiny_imagenet}, LSFormer achieves a top-1 classification accuracy of 71.61\% with a theoretical energy consumption of only 5.46 mJ, outperforming existing state-of-the-art SNN models. Specifically, compared to the methods on the VGG-13 architecture such as ASGL \cite{wang2023adaptive} (56.57\%) and LM-H \cite{hao2023progressive} (59.51\%), LSFormer achieves accuracy improvements of 15.04\% and 12.10\%, respectively.

When compared with Transformer-based SNN models such as Spikformer (66.03\%, 20.82 mJ) \cite{zhou2023spikformer}, LSFormer surpasses by 5.58\% with nearly four times lower energy consumption. The energy consumption of LSFormer (5.46 mJ) is 1.45 times that of SAFormer (2.23 mJ), but its accuracy (71.61\%) is 4.33\% higher than that of SAFormer (67.28\%). The higher accuracy gain per unit energy consumption makes LSFormer more competitive.

\subsection{Results and Analysis on Neuromorphic Datasets} \label{4.2}
As shown in Table~\ref{tab:performancen4}, LSFormer demonstrates competitive performance on neuromorphic datasets. On CIFAR10-DVS dataset, LSFormer achieves a top-1 classification accuracy of 84.30\% with 16 timesteps, significantly outperforming Spikformer \cite{zhou2023spikformer} and Spikingformer \cite{zhou2023spikingformer} by 3.40\% and 3.00\%, respectively.  

On DVS128 Gesture dataset, LSFormer achieves a top-1 classification accuracy of 98.60\% with 16 timesteps. The tdBN \cite{zheng2021going} achieves an accuracy of 96.90\% with 40 timesteps. By contrast, LSFormer not only improves top-1 classification accuracy by 1.70\% but also reduces the number of timesteps by 60\%.  Even with only 10 timesteps, LSFormer still achieves a top-1 classification accuracy of 97.90\%, improving 1.00\% than Spikformer \cite{zhou2023spikformer}. 

\begin{table}[t]
\setlength{\tabcolsep}{5pt} 
\centering
\small  
\caption{Experimental results  on CIFAR10-DVS and DVS128 Gesture. `Accuracy' denotes `Top-1 Classification Accuracy (\%)', `T' denotes `Timesteps'.}
\label{tab:performancen4}
\begin{tabular}{lcccc}
\toprule
\multirow{2}{*}{\textbf{Method}} & \multicolumn{2}{c}{\textbf{CIFAR10-DVS}} & \multicolumn{2}{c}{\textbf{DVS128 Gesture}} \\ \cmidrule(r){2-3} \cmidrule(r){4-5}
& \textbf{T} & \textbf{Accuracy} & \textbf{T} & \textbf{Accuracy} \\
\midrule
LIAF \cite{wu2021liaf} & 10 & 70.4 & 60 & 97.6 \\
tdBN \cite{zheng2021going} & 10 & 67.8 & 40 & 96.9 \\
PLIF \cite{fang2021incorporating} & 20 & 74.8 & 20 & 97.6 \\
Transformer \cite{10237117} & - & 71.2 & - & - \\
\midrule
\multirow{2}*{Spikformer \cite{zhou2023spikformer}} & 10 & 78.9 & 10 & 96.9 \\
                         & 16 & 80.9 & 16 & 98.3 \\
\midrule
\multirow{2}*{Spikingformer \cite{zhou2023spikingformer}} & 10 & 79.9 & 10 & 96.2 \\
                             & 16 & 81.3 & 16 & 98.3 \\
\midrule
S-Transformer \cite{yao2023spikedriven} & 16 & 80.0 & 16 & 99.3 \\
TIM \cite{shen2024tim} & 10 & 81.6 & - & - \\
SAFormer \cite{zhang2025combining} &16 & 81.3 & 16 & 98.3 \\
QKformer \cite{zhou2024qkformer} & 16 & 84.0 & 16 & 98.6 \\
\midrule
\textbf{LSFormer (Ours)} & \textbf{10} & \textbf{$\textbf{81.90}\pm \textbf{0.20}$} & \textbf{10} & \textbf{$\textbf{97.90}\pm \textbf{0.10}$} \\
                                    
                                             \textbf{LSFormer (Ours)} & \textbf{16} & \textbf{$\textbf{84.30}\pm \textbf{0.20}$} & \textbf{16} & \textbf{$\textbf{98.60}\pm \textbf{0.30}$} \\
\bottomrule
\end{tabular}
\end{table}

\begin{table}[h]
\centering
\setlength{\tabcolsep}{12pt}
\small
\caption{Experimental results on the N-CALTECH101 dataset. `Accuracy' denotes `Top-1 Classification Accuracy (\%)', and `T' denotes `Timesteps'.}
\label{tab:performancen101}
\begin{tabular}{lcc}
\toprule
\textbf{Method} & \textbf{T} & \textbf{Accuracy} \\
\midrule
ResNet-18 \cite{SHEN2023119170}        & 10 & 75.3 \\
Event Transformer \cite{10237117}     & -- & 78.9 \\
RMSNN \cite{2015Detection}            & 10 & 77.9 \\
TIM \cite{shen2024tim}                & 10 & 79.0 \\
\midrule
\textbf{LSFormer (Ours)}                       & \textbf{10} & \textbf{$\textbf{87.60} \pm \textbf{0.10}$} \\
\bottomrule
\end{tabular}
\end{table}
We further evaluate LSFormer on the more challenging N-CALTECH101 dataset. As shown in Table~\ref{tab:performancen101}, LSFormer achieves 87.60\% top-1 classification accuracy with 10 timesteps, and surpasses the TIM \cite{shen2024tim} (79.00\%) by 8.60\%.

Notably, LSFormer outperforms ResNet \cite{SHEN2023119170} (75.30\%) by 12.30\% and the ANN-based Transformer \cite{10237117} (78.90\%) by 8.70\%, respectively. These results further validate the superior performance and effectiveness of our LSFormer in neuromorphic classification tasks.

\subsection{Attention Visualization}\label{4.4}
\begin{figure}[h]
\centering 
\includegraphics[width=1\linewidth]{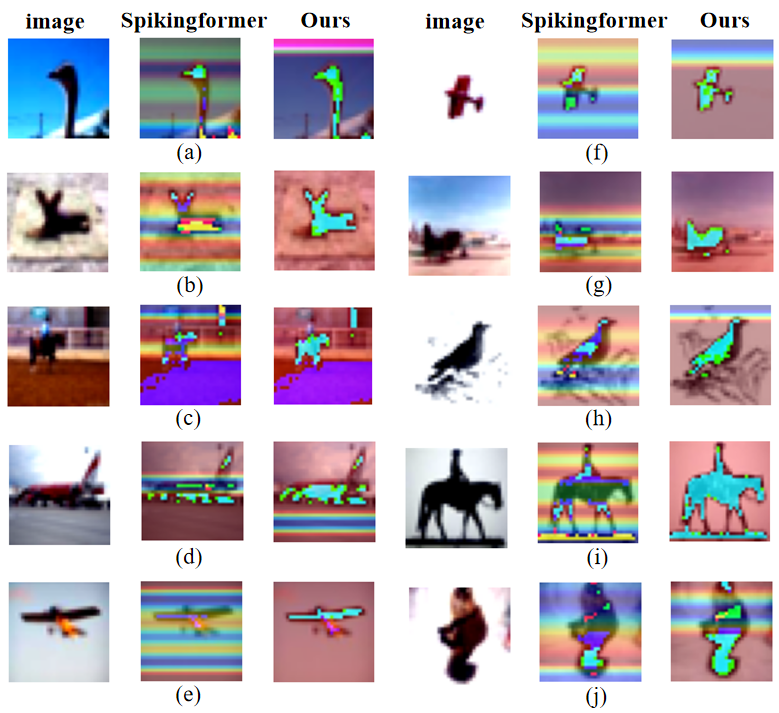}
\caption{\label{fig:Grad-CAM} Grad-CAM visualization of the last block of our LSFormer and Spikingformer \cite{zhou2023spikingformer} on the CIFAR-10 dataset. (a)-(j) are demonstrations of 10 random samples extracted from the dataset.}
\end{figure}

To further assess the recognition performance of our proposed LSFormer, we employ Grad-CAM \cite{jacobgilpytorchcam} to visualize the attention maps generated by the last encoder block in Spikingformer \cite{zhou2023spikingformer} and our LSFormer. As illustrated in Fig.~\ref{fig:Grad-CAM}, compared with Spikingformer, LSFormer produces more continuous and concentrated attention regions, reflecting that LSFormer exhibits a stronger ability to focus on task-relevant features.

Notably, Fig.~\ref{fig:Grad-CAM} (a) exhibits spatial patterns that are predominantly distributed along the vertical direction, whereas Fig.~\ref{fig:Grad-CAM} (d) shows the horizontal distribution. In LSFormer, attention is computed independently along the horizontal and vertical directions, which enables more effective capture of such spatial distribution patterns. Furthermore, LSFormer demonstrates superior recognition capability for objects of the same category but with different scales. For instance, the horse instances (Fig.~\ref{fig:Grad-CAM} (c) and Fig.~\ref{fig:Grad-CAM} (i)) differ significantly in spatial scale and posture, while LSFormer can still precisely focus on their critical features, reflecting its stronger multi-scale feature extraction capability.

\subsection{Ablation Study}\label{4.3}
We conduct ablation studies on the LS-SSA and SPooling to analyze their specific impacts on model performance. Ablation experiments are performed on both static image datasets and the neuromorphic datasets.

\subsubsection{Spiking Response Pooling} 

To validate the effectiveness of SPooling, we replace the max pooling in Spikingformer \cite{zhou2023spikingformer} (baseline) with SPooling, and maintain all other experimental settings unchanged. As shown in Table~\ref{tab:dilation_impact}, SPooling improves the top-1 classification accuracy of Spikingformer by 1.40\% and 1.79\% on CIFAR10-DVS and CIFAR-100 datasets, respectively. The experiments thoroughly demonstrate the superiority of SPooling compared with the traditional max pooling in spike-based architectures.  We additionally compare SPooling with other existing improved pooling strategies. Specifically, we replace the original max pooling in the baseline with adaptive pooling \cite{lin2013network}. The results show that adaptive pooling yields inferior performance on both CIFAR10-DVS and CIFAR-100 datasets compared to SPooling, and even causes a performance drop on CIFAR-100 and CIFAR10-DVS datasets compared with the baseline.

To further evaluate the specific contribution of SPooling in enhancing informative feature representation, we conduct a systematic analysis of the threshold parameter \(\lambda\)  (covering both extreme and intermediate values) on the CIFAR-100 and CIFAR10-DVS datasets.

\begin{table}[h]
\centering
\caption{Ablation studies of SPooling and LS-SSA on CIFAR10-DVS and CIFAR-100 datasets. 
The baseline is Spikingformer \cite{zhou2023spikingformer}. 
`Accuracy' denotes `Top-1 Classification Accuracy (\%)'. * indicates the results from three runs of our own implementation on baseline.}
\label{tab:dilation_impact}
\small
\setlength{\tabcolsep}{4pt}  
\begin{tabular}{lcc}
\hline
\multirow{2}{*}{\textbf{Model}} & \multicolumn{2}{c}{\textbf{Accuracy}} \\
\cmidrule(r){2-3}
 & \textbf{CIFAR10-DVS} & \textbf{CIFAR-100} \\
\hline
Baseline*  & \textbf{$81.20\pm 0.10$}  & \textbf{$79.23 \pm 0.03$} \\

Baseline w/ adaptive pooling \cite{lin2013network} & \textbf{$80.90\pm 0.26$} & \textbf{$79.05\pm 0.06$} \\
Baseline w/ SPooling & $82.60\pm 0.22$ & \textbf{$81.02 \pm 0.06$} \\
Baseline w/ MSDA \cite{jiao2023dilateformer} &  \textbf{$81.80 \pm0.20$} & \textbf{$80.12\pm0.02$} \\
Baseline w/ LS-SSA & $83.20\pm 0.15$& \textbf{$81.58 \pm 0.05$}  \\
\mbox{Baseline w/ SPooling, LS-SSA} &\textbf{$84.30\pm0.20$} & \textbf{$82.00\pm0.04$} \\
\hline
\end{tabular}
\end{table}

\begin{table}[t]
\setlength{\tabcolsep}{14pt}  
\centering
\small
\caption{Impact of different threshold parameters \(\lambda\) in SPooling on CIFAR10-DVS and CIFAR-100 datasets. `Accuracy' denotes `Top-1 Classification Accuracy(\%)'.}
\label{tab:tau_impact}
\begin{tabular}{lcc}
\toprule
\multirow{2}{*}{\textbf{$\lambda$}} & \multicolumn{2}{c}{\textbf{Accuracy }} \\
\cmidrule(r){2-3}
& \textbf{CIFAR10-DVS} & \textbf{CIFAR-100} \\
\midrule
0.0 & $82.10\pm 0.17$  & $79.24\pm 0.16$  \\
0.1 & $82.03\pm 0.12$  & $81.39\pm 0.03$  \\
0.2 & $83.07\pm 0.21$ & $81.56\pm 0.14$  \\
\textbf{0.3} & \textbf{$\textbf{84.30} \pm\textbf{ 0.20}$}  & \textbf{$\textbf{82.00} \pm \textbf{0.04}$} \\
0.4 & $83.73\pm 0.15$ & $81.70\pm 0.07$ \\
0.5 & $82.47\pm 0.32$  & $81.30\pm 0.02$ \\
0.6 & $83.90 \pm 0.36$  & $81.66\pm 0.05$  \\
0.7 & $83.00\pm 0.20$  & $81.09\pm 0.08$  \\
0.8 & $82.47\pm 0.15$  & $81.37\pm 0.06$  \\
0.9 & $81.63\pm 0.15$  & $80.43\pm 0.08$  \\
1.0 & $82.57\pm 0.25$  & $79.29 \pm 0.10$ \\
\bottomrule
\end{tabular}
\end{table}

We have conducted a comprehensive grid search over the experimental range of $\lambda$ from 0.0 to 1.0 with a step size of 0.1. The choice of a 0.1 step size ensures that a sufficient number of representative threshold values are tested within a reasonable experimental duration, while also providing enough granularity to assess performance variations. As shown in Table~\ref{tab:tau_impact}, the best performance within the experimental range is achieved at \(\lambda = 0.3\) (84.30\% on CIFAR10-DVS dataset and 82.00\% on CIFAR-100 dataset). Notably, when \(\lambda = 0.0\), the average pooling results in practice are typically non-negative and naturally meet or exceed the threshold $\lambda$. This means no average pooling results are filtered out, and SPooling degenerates into a weighted combination of max pooling and average pooling (hybrid pooling \cite{yu2014mixed}), thereby disabling the ability of SPooling to selectively emphasize sub-threshold responses. In contrast, when \(\lambda = 1.0\), most average pooling results are filtered out. As shown in Table~\ref{tab:tau_impact}, the performance degradation at these two extreme values (\(\lambda = 0.0\) and \(\lambda = 1.0\)), highlights the essential role of \(\lambda\) as a balancing parameter, which facilitates more effective preservation of salient features. Except for the two extreme values exhibiting degraded performance, SPooling maintains consistently stable performance at all other values within the experimental range. This validates the effectiveness and robustness of our proposed SPooling.
 
\subsubsection{Local Structure-Aware Spiking Self-Attention} 

To verify the effectiveness of our proposed LS-SSA, we replace the original SSA in Spikingformer with LS-SSA. As shown in Table~\ref{tab:dilation_impact}, the variant model (with LS-SSA) outperforms the baseline Spikingformer \cite{zhou2023spikingformer} (with SSA) by 2.00\% and 2.35\% in top-1 classification accuracy on CIFAR10-DVS and CIFAR-100 datasets, respectively.  Moreover, we also compare LS-SSA with a representative multi-scale dilated attention mechanism (MSDA \cite{jiao2023dilateformer}). The results demonstrate that LS-SSA consistently achieves superior performance, outperforming MSDA by 1.40\% on CIFAR10-DVS and 1.46\% on CIFAR-100. These results clearly validate the effectiveness of our proposed LS-SSA.

To further verify the necessity of each component of LS-SSA, we conduct a comprehensive ablation study by individually disabling SDA, CRA, SCDF, and DWC embedded in SDA. As shown in Table~\ref{tab:ls_ssa_ablation}, removing any of these components consistently leads to performance degradation on both CIFAR10-DVS and CIFAR-100, indicating that each module makes a complementary and non-redundant contribution.
Specifically, removing SDA significantly weakens local structural modeling, while disabling CRA reduces the effectiveness of channel-wise modulation.
Moreover, removing the decoupled fusion mechanism and adopting a coupled design leads to inferior performance. Without the learnable modulation parameters $\alpha_m$ and $\beta_m$, spatial attention and channel attention are forced to share identical importance weights, preventing flexible and independent regulation of spatial structures and channel-wise modulation.

Notably, as shown in Table~\ref{tab:ls_ssa_ablation}, removing DWC in SDA also leads to a significant performance drop.
This demonstrates that DWC plays an important role in enhancing local feature integration and compensating for the sparsity introduced by dilated sampling, without introducing additional interference. We further provide visualization results of boundary region feature extraction before and after applying DWC in SDA in the Appendix~\ref{DWC}, demonstrating the effect of DWC on boundary feature enhancement.

\begin{table}[t]
\centering
\small
\caption{Ablation study of different components in LS-SSA on CIFAR10-DVS and CIFAR-100. `Accuracy' denotes `Top-1 Classification Accuracy (\%)'.}
\label{tab:ls_ssa_ablation}
\begin{tabular}{lcc}
\hline
\toprule
\multirow{2}{*}{\textbf{Model Configuration}} & \multicolumn{2}{c}{\textbf{Accuracy }} \\
\cmidrule(r){2-3}
& \textbf{CIFAR10-DVS} & \textbf{CIFAR-100} \\
\midrule
LS-SSA                          & $84.30\pm 0.20$ & $82.00\pm 0.04$ \\
LS-SSA w/o SDA                         &$82.67\pm 0.32$ & $80.29\pm 0.02$  \\
LS-SSA w/o CRA                         & $82.80\pm 0.20$ & $81.37\pm 0.03$\\
LS-SSA w/o SCDF                        & $83.53\pm 0.21$ & $81.48\pm 0.04$\\
SDA w/o DWC                 & $83.00\pm 0.20$ &$81.44\pm 0.04$ \\
\hline
\end{tabular}

\end{table}

\subsubsection{The Influence of Dilation Parameters} \label{appx:c4}

We analyze the impact of different dilation scales in LS-SSA on classification performance using the CIFAR10-DVS and CIFAR-100 datasets. Unlike Spiking Self-Attention \cite{zhou2023spikingformer}, which computes attention globally, LS-SSA operates within local dilated windows to model spatial relationships. As shown in Table~\ref{tab:dilation_impact_2}, we select several representative dilation-rate combinations (covering both multi-scale and single-scale settings) to analyze their effects on performance. LS-SSA demonstrates robustness to variations in dilation rates, with only minor changes in accuracy across different configurations.
\begin{table}[h]
\small
\centering
\caption{Comparative impact of dilation parameters on classification accuracy for CIFAR10-DVS and CIFAR-100 datasets. `Accuracy' denotes `Top-1 Classification Accuracy (\%)'. Statistical significance (independent samples t-test) is marked as: *$p \textless 0.05$, **$p \textless 0.01$, compared to the `Multi-[1,2,3]' group.}
\label{tab:dilation_impact_2}
\begin{tabular}{lcll}
\hline
\multirow{2}{*}{\textbf{Scale}} & \multirow{2}{*}{\textbf{Dilation Rate}} & \multicolumn{2}{c}{\textbf{Accuracy}} \\ 
\cmidrule(r){3-4}
        & &  \textbf{CIFAR10-DVS} & \textbf{CIFAR-100} \\
\hline
 \multirow{4}*{Multi-}
 & [1,2,3] & $84.30\pm 0.20$ & $82.00\pm 0.04$ \\  
 & [2,3,4] & $82.83\pm 0.15^{**}$ & $81.43\pm 0.03^{**}$\\
 & [1,2,4] & $82.63\pm 0.12^{**}$ & $81.67\pm 0.02^{**}$\\ 

 & [1,2,3,4] & $82.70\pm 0.44^{*}$ & $81.79\pm 0.03^{**}$ \\

\hline
 \multirow{3}*{Single-} & [1] & $82.83\pm 0.21^{**}$ & $81.50\pm 0.03^{**}$ \\
 & [2] & $83.20\pm 0.20^{*}$ & $81.55\pm 0.02^{**}$\\
 & [3] & $82.97\pm 0.21^{**}$ & $81.59 \pm 0.04^{**}$\\ 
\hline
\end{tabular}
\end{table}

Multi-scale dilated attention, particularly with dilation rates [1,2,3], significantly outperforms all single-scale configurations (i.e., [1], [2], and [3]) with p < 0.01 on both datasets. This indicates that multi-scale dilated attention can provide more comprehensive information than single-scale dilated attention. As shown in Table~\ref{tab:dilation_impact_2}, including a dilation rate of 4 (e.g., [1,2,4], [2,3,4], and [1,2,3,4]) leads to a decrease in top-1 classification accuracy compared to [1,2,3]. For CIFAR10-DVS, accuracy drops by 1.47\% ([2,3,4]), 1.67\% ([1,2,4]), and 1.60\% ([1,2,3,4]). For CIFAR-100, accuracy decreases by 0.57\% ([2,3,4]), 0.33\% ([1,2,4]), and 0.21\% ([1,2,3,4]).
This decrease occurs because dilation rates larger than 3 create overly sparse sampling within local windows, causing most neighboring information to be neglected. Given that the feature map resolution is only $8\times8$, a dilation rate of 4 yields receptive fields of $1\times9$ and $9\times1$ along the horizontal and vertical directions, respectively, which exceed the feature map size. This produces excessively sparse sampling and introduces irrelevant spatial noise, reducing the ability of the model to learn discriminative local features. Therefore, in our experiments, we limit the maximum dilation rate to 3. These results suggest that LS-SSA achieves the best trade-off between performance and efficiency with multi-scale dilation while avoiding excessively large dilation rates.

\subsubsection{The Influence of Different Timesteps on CIFAR-10 and CIFAR-100 Datasets} 

\begin{figure}[h]
\centering
\includegraphics[width=1\linewidth]{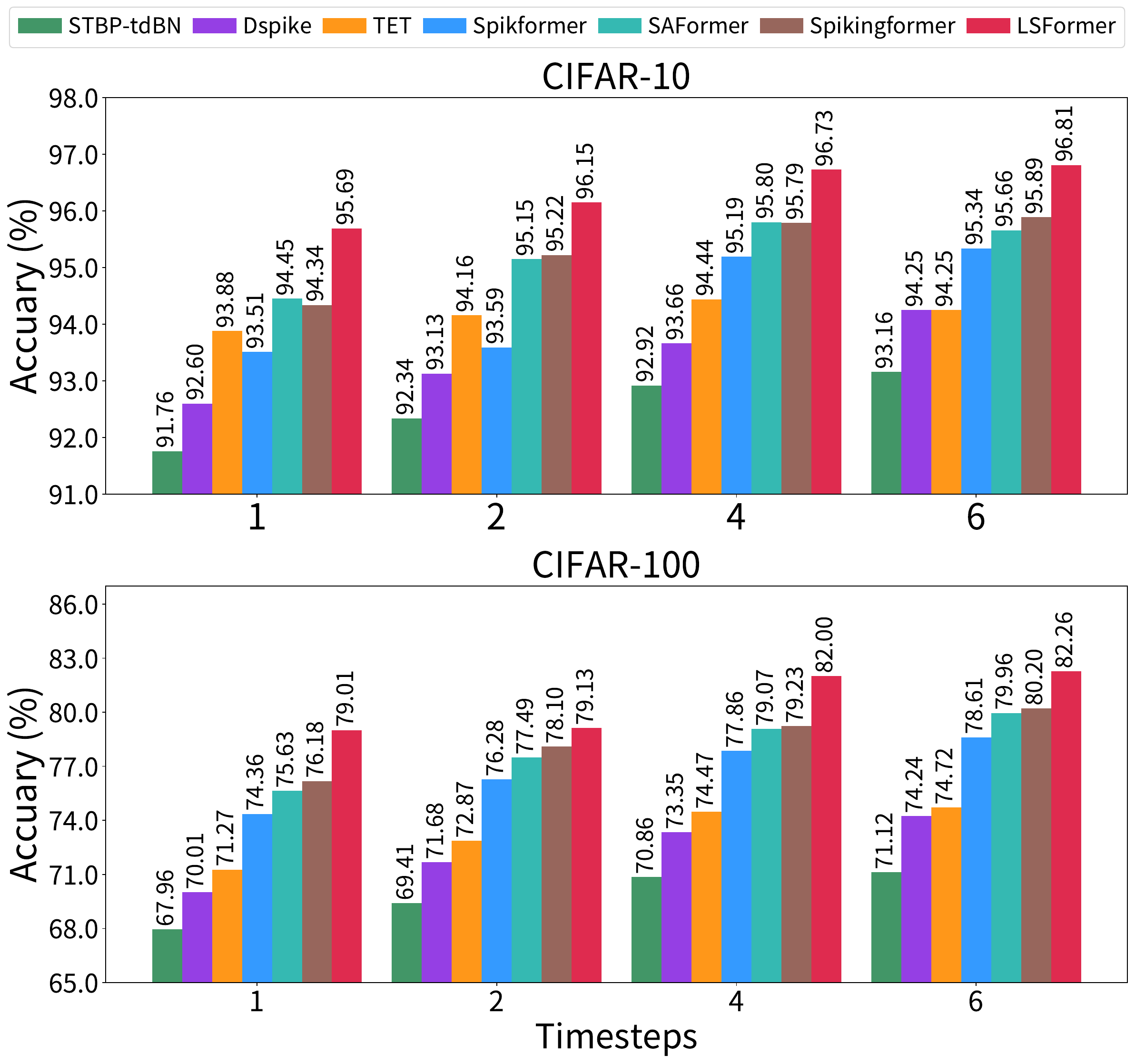}
\caption{\label{fig:timesteps}Performance comparison of different models under varying timesteps on CIFAR-10 and CIFAR-100 datasets. Results are reported with deviation ranges calculated over three independent runs.}
\end{figure}

The experimental results for different simulation timesteps are shown in Fig.~\ref{fig:timesteps}. 

As the number of timesteps increases, all models show consistent performance improvements on CIFAR-10 and CIFAR-100 datasets, indicating that these models are already capable of effectively modeling temporal information.  
Across all timesteps, LSFormer consistently outperforms the other models.  In SNNs, the membrane potentials of neurons accumulate input spikes over time, naturally integrating information across timesteps. 
As a result, LSFormer can exploit temporal information without the need for an explicit temporal attention mechanism, thus achieving high classification performance without incurring additional computational overhead. 
Importantly, as shown in Table~\ref{tab:performancen4}, this behavior is not limited to CIFAR-10 and CIFAR-100 datasets, and similar improvements are observed on neuromorphic datasets.

\section{Conclusion}
In this paper, we propose a novel Local Structure-Aware Spiking Self-Attention (LS-SSA) mechanism, which computes self-attention within the local dilated windows, effectively capturing multi-scale features. Furthermore, we introduce Spiking Response Pooling (SPooling), designed to extract richer and more informative feature representations. Extensive experiments on both static and neuromorphic datasets demonstrate that our approach consistently achieves state-of-the-art performance. 
Importantly, both LS-SSA and SPooling are derived from the intrinsic characteristics of spike-based computation, rather than generic architectural enhancements, making them particularly suitable for directly trained Transformer-based SNNs.
Notably, we pioneer the validation of local dilated window attention in directly trained Transformer-based SNNs. In the future, we plan to explore specific hardware optimizations and conduct benchmarking on real platforms to comprehensively assess the practical potential of our methods in real-world deployment scenarios.

\section{Acknowledgment}
This work was in part supported by the Brain Science and Brain-like Intelligence Technology - National Science and Technology Major Project (Grant No. 2025ZD0215600), the National Natural Science Foundation of China (Grant No. 92370103), and in part by the Xiaomi Foundation.
\appendix

\subsection{Dataset Details}\label{Dataset}
CIFAR-10 \cite {alex2009learning} and CIFAR-100 \cite {alex2009learning} are widely used for image classification tasks. Both datasets contain 50,000 training images and 10,000 test images, each with a resolution of 32 $\times$ 32 pixels. Specifically, CIFAR-10 consists of 10 distinct categories, while CIFAR-100 comprises 100 categories. Tiny-ImageNet-200 \cite {alex2009learning} is a subset of the ImageNet dataset \cite {alex2009learning}, consisting of 200 categories. Each category contains 500 training images, 50 validation images, and 50 test images, with a reduced resolution of 64 $\times$ 64 pixels. 

The N-CALTECH101 dataset \cite {orchard2015converting} is transformed from the Caltech101 dataset \cite {2004Learning}, providing a time-based visual event representation for the original static image categories. CIFAR10-DVS \cite {li2017cifar10} converts CIFAR-10 images into 10,000 event streams using Dynamic Vision Sensors (DVS), with an original resolution of 128 $\times$ 128 pixels and 10 categories. The DVS 128 Gesture dataset \cite {amir2017low} includes 11 distinct gesture categories, with 1,176 training samples and 288 test samples in total.

All experiments are conducted on a cluster of NVIDIA RTX 4090 GPUs. The implementation is based on the leaky integrate-and-fire (LIF) neuron, PyTorch \cite{NEURIPS2019_bdbca288}, SpikingJelly \cite{fang2023spikingjellyopensourcemachinelearning}, and the PyTorch image models library (timm) \cite{rw2019timm}. For a fair comparison with existing works, we adopt the data augmentation and training strategies proposed in \cite{zhou2023spikingformer}. For static image datasets, the patch size is set to 8 $\times$ 8. The LSFormer architecture consists of four encoder blocks with a feature embedding dimension of 384. We employ the AdamW optimizer for 450 training epochs, including an initial 20 warm-up epochs and a final 10 cool-down epochs. A cosine decay learning rate scheduler is adopted, where the learning rate is decayed by a factor of 10 every 30 epochs. The base learning rate is set to 0.0005, the minimum learning rate to 0.00001, the batch size is set to 64, and the weight decay is set to 0.06. 

For neuromorphic datasets, the patch embedding dimension is 256, and the patch size is 8 $\times$ 8. The LSFormer architecture comprises two encoder blocks. The number of timesteps is set to 10 or 16. The AdamW optimizer is used, with 100 training epochs for N-CALTECH101 and CIFAR10-DVS, and 200 epochs for DVS 128 Gesture. Each training process includes 10 warm-up epochs and 10 cool-down epochs. A cosine decay learning rate scheduler is adopted, with the learning rate decayed by a factor of 10 every 20 epochs. The base learning rate is 0.001, the batch size is 8, and the weight decay is 0.06.

\subsection{Computational Complexity Analysis}\label{Computational}
Let $N = H \times W$ denote the number of spatial tokens and $D$ denote the number of channels. For  VSA \cite{10.5555/3295222.3295349} and SSA \cite{zhou2023spikformer,zhou2023spikingformer}, attention is computed globally across all spatial tokens, resulting in a quadratic computational cost.
Specifically, as shown in Table~\ref{tab:complexity}, both methods exhibit a time complexity of $\mathcal{O}(N^{2}D)$ and a space complexity of $\mathcal{O}(N^{2} + ND)$.

\begin{table}[h]
\small
\centering
\caption{Computational complexity comparison of different attention mechanisms.}
\label{tab:complexity}
\begin{tabular}{lcc}
\toprule
\textbf{Method} & \textbf{Time Complexity} & \textbf{Space Complexity} \\
\midrule
VSA  \cite{10.5555/3295222.3295349}  & $\mathcal{O}(N^{2}D)$ & $\mathcal{O}(N^{2} + ND)$ \\
SSA \cite{zhou2023spikformer,zhou2023spikingformer}  & $\mathcal{O}(N^{2}D)$ & $\mathcal{O}(N^{2} + ND)$ \\
SDSA \cite{yao2023spikedriven} & $\mathcal{O}(ND)$     & $\mathcal{O}(ND)$ \\
\textbf{LS-SSA} & $\mathcal{O}(ND)$ & $\mathcal{O}(ND)$ \\
\bottomrule
\end{tabular}
\end{table}

The spatial attention in LS-SSA is computed within local windows of size $1 \times k$ or $k \times 1$, where $k$ is a resolution-independent constant satisfying $k \ll N$. Therefore, its contribution can be treated as a constant factor and does not affect the asymptotic complexity.  As a result, LS-SSA achieves an overall time complexity of $\mathcal{O}(ND)$ and a space complexity of $\mathcal{O}(ND)$. LS-SSA completely eliminates the quadratic dependency on $N$, significantly improving computational efficiency.     This property makes LS-SSA particularly suitable for SNNs, where feature maps are sparse and efficient processing is critical.

\begin{figure}[h]
\centering
\includegraphics[width=0.7\linewidth]{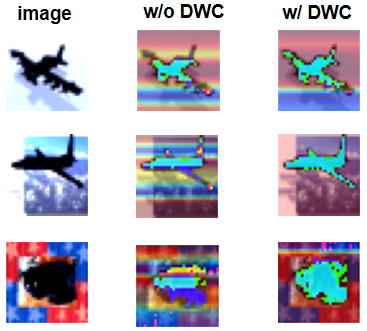}
\caption{Grad-CAM visualizations of the last block in LSFormer, comparing the SDA module with and without DWC on randomly selected samples from the CIFAR-10 dataset.}
\label{fig:DWC-CAM}
\end{figure}

\subsection{Visualization Analysis of DWC in SDA}\label{DWC}
During the sampling of boundary query slices in SDA, zero-padding is applied, which may lead to incomplete feature extraction near edge regions. 
To alleviate this issue, DWC is introduced to complement the corresponding value matrix slices.

As shown in Fig.~\ref{fig:DWC-CAM}, Grad-CAM visualizations from the last block of LSFormer compare the SDA module with and without DWC, providing an intuitive analysis of the effect of DWC.
Without DWC, the attention  maps exhibit weakened and discontinuous responses around object boundaries, indicating insufficient feature coverage in these regions. 
In contrast, incorporating DWC produces more continuous and complete attention, reflecting enhanced feature representation near boundary areas.  These visual results qualitatively verify the effectiveness of DWC in improving boundary-related features of objects in SDA.

\subsection{The Theoretical Energy Consumption Compared with Transformer-based SNNs and ANNs}\label{SOPs}

Table~\ref{tab:energy_sops} illustrates the energy consumption, synaptic operations (SOPs), and classification accuracy on CIFAR-10 dataset. Compared to Spikformer \cite{zhou2023spikformer}, LSFormer reduces energy consumption by 73.0\% while achieving 1.22\% higher accuracy. When compared to SEW-ResNet \cite{fang2021deep}, the representative of ANNs, LSFormer reduces energy consumption by 84.3\%, while achieving 6.71\% higher accuracy.

\begin{table}[h]
\centering
\small
\setlength{\tabcolsep}{4pt} 
\caption{The theoretical energy consumption compared with Transformer-based SNNs and ANNs. `Accuracy' denotes `Top-1 Classification Accuracy (\%)'.}
\begin{tabular}{lccc}
\toprule
\textbf{Method} & \textbf{SOPs (G)} & \textbf{Power (mJ)} & \textbf{Accuracy} \\
\midrule
SEW-ResNet \cite{fang2021deep} & -- &  8.88 & 90.02 \\
Spikformer \cite{zhou2023spikformer} & 66.88 &  5.15 & 95.51 \\
Spikingformer \cite{zhou2023spikingformer} &   14.81 &   1.14 & 95.81 \\
\textbf{LSFormer (Ours)} &   \textbf{18.05} &  \textbf{1.39} & \textbf{96.73} \\
\bottomrule
\end{tabular}
\label{tab:energy_sops}
\end{table}

Although LSFormer consumes 0.25 mJ more energy than Spikingformer \cite{zhou2023spikingformer}, it achieves a 0.92\% improvement in top-1 classification accuracy. These results highlight the balanced performance-energy trade-off of LSFormer, demonstrating its ability to effectively leverage informative spike activities for improved recognition capability without incurring excessive energy overhead.

\begin{figure}[h]\label{Firing}                                                                    
\centering 
\includegraphics[width=0.8\linewidth]{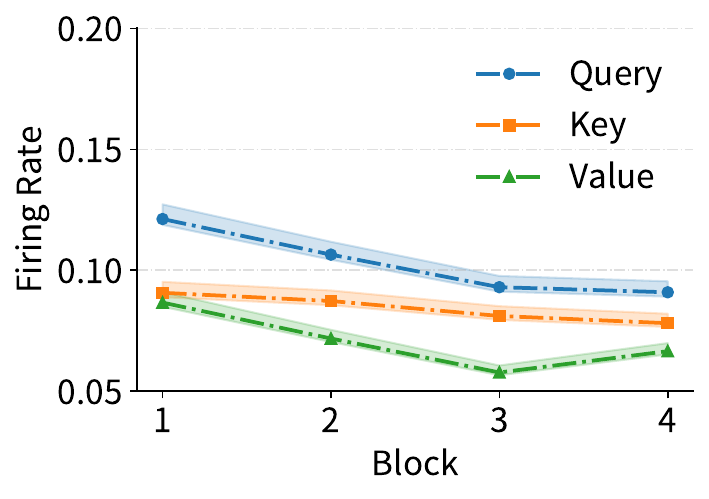}
\caption{\label{fig:Firing} Firing rate of query, key and value of blocks in LSFormer-4-384 on CIFAR-10 dataset.
}
\end{figure}

\subsection{Firing Rate of Query, Key and Value}\label{Firing1}
As shown in Fig.~\ref{fig:Firing}, the query, key, and value in LS-SSA exhibit sparse spiking activity, which is well suited for efficient spike-driven computation. The firing rates evolve smoothly across encoder blocks, indicating stable information propagation across encoder blocks.

\bibliography{reference}
\bibliographystyle{IEEEtran}

\newpage

\vspace{11pt}

\vfill

\end{document}